\documentclass[runningheads]{llncs}

\usepackage{graphicx}
\usepackage{comment}
\usepackage{amsmath,amssymb} 
\usepackage{color}

\usepackage[section]{placeins}
\usepackage{filecontents}
\usepackage{subcaption}
\usepackage{siunitx}
\usepackage{bm}
\usepackage{multirow}
\usepackage{csquotes}

\newcommand{\etal}{\textit{et al.}}

\usepackage[breaklinks=true,bookmarks=false]{hyperref}
\usepackage{xcolor}

\begin{document}
\pagestyle{headings}
\mainmatter

\title{Unconstrained Text Detection in Manga: a New Dataset and Baseline} 

\titlerunning{ }
%
\author{Juli\'an Del Gobbo\inst{1} \and
Rosana Matuk Herrera\inst{2}}
\authorrunning{ }
%
\institute{Departamento de Computaci\'on, FCEN, 
Universidad de Buenos Aires, Argentina
\and Departamento de Ciencias B\'asicas, Universidad Nacional de Luj\'an, Argentina
}
\maketitle

\begin{abstract}

The detection and recognition of unconstrained text is an open problem in research.  
Text in comic books has unusual styles that raise many challenges for text detection.  
This work aims to binarize text 
in a comic genre with highly sophisticated text styles: Japanese manga. To overcome the lack of a manga dataset with text annotations at a pixel level,  
we create our own. To improve the evaluation and search of an optimal model, in addition to standard metrics in binarization, we implement other special metrics.
Using these resources, we designed and evaluated a deep network model, outperforming current methods for text binarization in manga in most metrics.

\end{abstract}

\section{Introduction}

\begin{figure}[h!]
\centering
\begin{subfigure}[b]{0.73\textwidth}
\includegraphics[width=\textwidth]{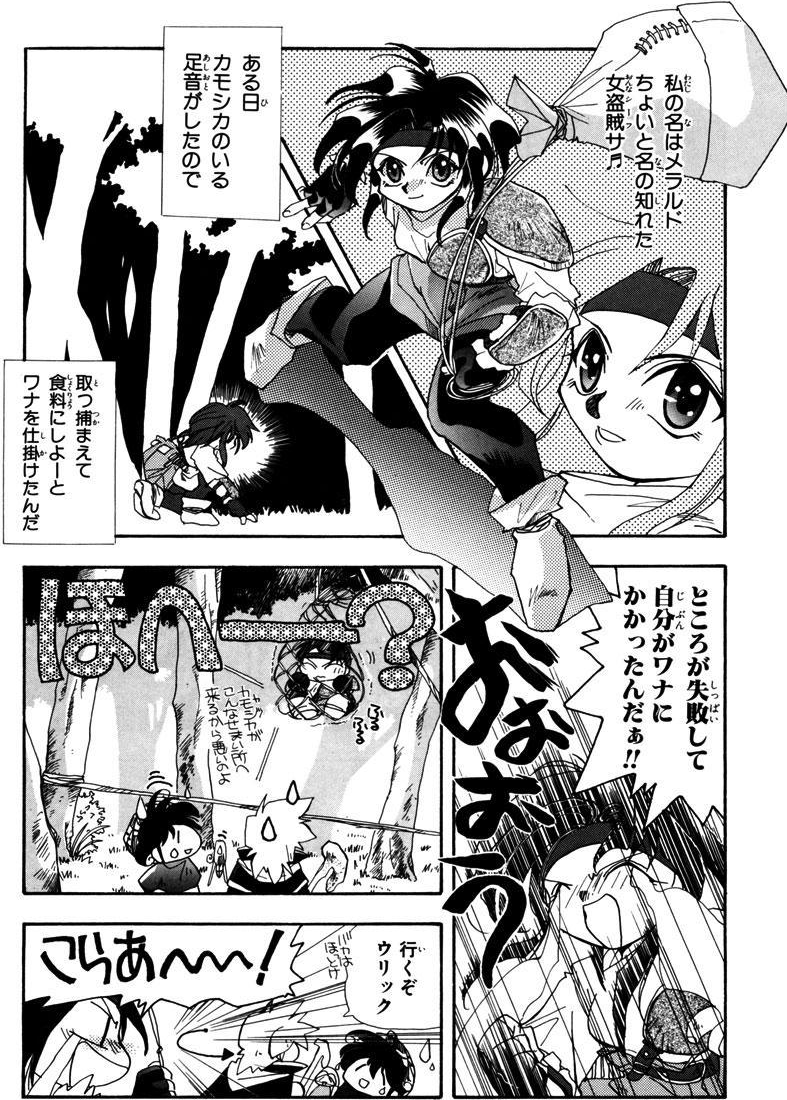}
\caption{}
\label{manga1}
\end{subfigure}
\begin{subfigure}[b]{0.24\textwidth}
\includegraphics[width=\textwidth]{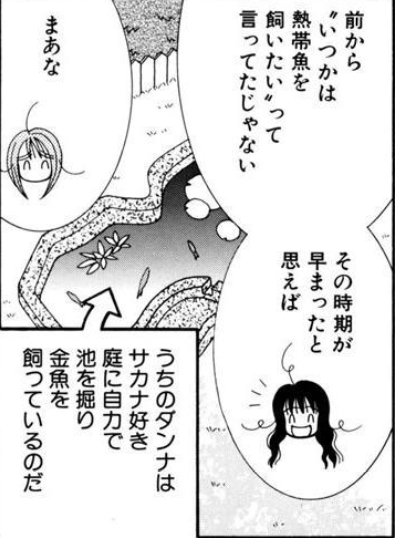} 
\caption{}
\label{manga2}
\includegraphics[width=\textwidth]{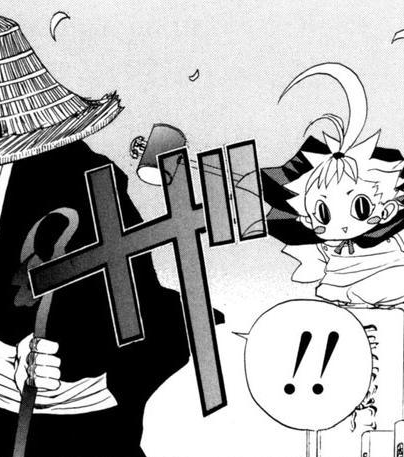}
\caption{}
\label{manga3}
\includegraphics[width=\textwidth]{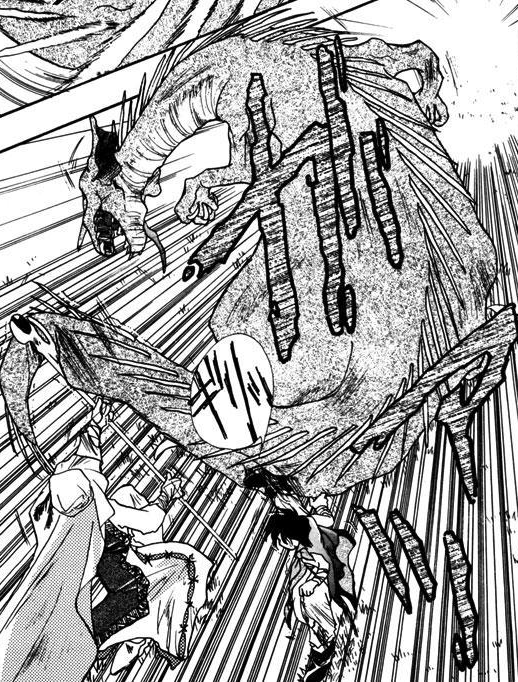}
\caption{}      
\label{manga4}
\end{subfigure}
\caption{
Pictures showing the diversity of text styles in manga. (a) The dialogue balloons could have unconstrained shapes and border styles. The text could have any style and fill pattern, and could be written inside or outside the speech balloons. Note also that the frames could have non-rectangular shapes, and the same character could be in multiple frames. (b) Example of manga extract featuring non-text inside speech bubbles. (c) The same text character can have diverse levels of transparency. (d) Text characters could have a fill pattern similar to the background.
All images were extracted from the \texttt{Manga109} dataset \cite{manga109}\cite{Matsui_2016}\cite{ogawa2018object}:
(a) and (d) ``Revery Earth'' \textcopyright {Miyuki Yama}, (b) ``Everyday Oasakana-chan'' \textcopyright {Kuniki Yuka}, (c) ``Akkera Kanjinchou'' \textcopyright {Kobayashi Yuki} 
}
\label{img:heads}
\end{figure}

 Manga is a type of Japanese comic book with a huge diversity of text and balloons styles in unconstrained positions (Fig. \ref{img:heads}). Japanese is a highly complex language, with three different alphabets and thousands of text characters. It also has about 1200 different onomatopoeia, which frequently appear in manga. Furthermore, characters often look very similar to the art in which they are embedded. These complexities make a text detection method for manga challenging to design.

A translation process of a manga picture consists of detecting the text, erasing it, inpainting the image, and writing the translated text on the image. As it is an intricate process, the translation  is usually done manually in manga, and only the most popular mangas are translated. The complexity of the Japanese language hinders the diffusion of manga, and other Japanese artworks, outside of Japan.  Automating the translation would lead to solving the linguistic barrier.
In this work, we focus on the first step of the translation process: text detection. To achieve high-quality inpainting of the image, the text detection should be made at a pixel level.

On one side, many works in text detection in comics have taken a balloon detection approach. However, in manga, the text and balloons are also part of the artwork. Thus, balloons could have a multiplicity of shapes and styles. Besides, the text can be outside the dialogue balloons (Figs. \ref{manga1},\ref{manga3},\ref{manga4}), or inside the balloon there could be non-text contents (Fig. \ref{manga2}), making a balloon detection approach unsuitable for this task. On the other side, most previous works in text detection have taken a box detection approach. However, manga contains texts that are deformed, extremely large, or are drawn on the cartoon characters, which are hard to identify with a single bounding box (Figs. \ref{manga1},\ref{manga3},\ref{manga4}).
Thus, 
we decide to make text segmentation at a pixel level, identifying pixels as either text or background.

This work aims  to detect unconstrained text characters in manga at a pixel level. The main contributions of this work  are the following:
\begin{itemize}
  \item The implementation and public release of many metrics for text detection at a pixel level, which others may reutilize, providing an easy way to find better models and compare results (Section \ref{metrics}).
  \item The creation and public release of a new dataset on text segmentation at a pixel level of Japanese manga  (Section \ref{dataset}). To the best of our knowledge, this is the first dataset of these characteristics in manga.
    \item 
    A simple and efficient model based on the U-net architecture  \cite{unet} that out-performs state-of-the-art works on text detection in Japanese manga, with good baseline metrics for future works
   (Sections \ref{model} and \ref{experiments}).
\end{itemize}

Despite the focus in manga, many techniques proposed in this work are general and  
could be used to do unconstrained text detection 
in other contexts.

\section{Related Work}\label{relatedwork}

\hspace{0.4cm} \textbf{Speech balloon detection} Several works have studied speech balloon detection in comics \cite{rigaud,rigaud_2019,liu_extraction,Dubray}. While this could be used to detect speech balloons and then consider its insides as text, the problem is that text in manga is not always inside speech balloons. Furthermore, there are a few cases where not everything inside the balloon is text (Fig. \ref{manga2}).

\textbf{Bounding box detection} Other works in text detection in manga, such as Ogawa \etal \cite{ogawa2018object} and Yanagisawa \etal \cite{Yanagisawa}, have focused on text bounding box detection of multiple objects, including text. Wei-Ta Chu and Chih-Chi Yu have also worked on bounding box detection of text \cite{chu}. 
Without restricting to manga or comics, there are many works every year that keep improving either bounding box or polygon text detection, one of the most recent ones being  Wang \etal  \cite{wang2019shape}. 
However, methods trained with rigid word-level bounding boxes exhibit limitations in representing the text region for unconstrained texts. Recently, Baek \etal  proposed a 
method (CRAFT) \cite{baek2019character} to detect unconstrained text in scene images. By exploring each character and affinity between characters, they generate non-rigid word-level bounding boxes.

\textbf{Pixel-level text segmentation} 
There is a long history of text segmentation \cite{text_segmentation_review} and image binarization \cite{Fletcher1988ARA}\cite{findingtextinimages}\cite{dibco_2009}\cite{dibco_2019} in the document analysis community related to historical manuscripts \cite{historical_manuscript_binarization}, maps \cite{map_binarization}, handwritten text \cite{Solihin1999IntegralRA}, documents \cite{document_binarization} and more. One of such works that does pixel level segmentation of text in document images is BCDU-net \cite{BCDUnet}. However,
in these works most of the image is text along with a few lines or figures, and the text is more simple than one of manga, which features a lot more context, wide variety of shapes and styles. 
Outside the document binarization community, there are very few works that do pixel-level segmentation of characters, as there are few datasets available with pixel-level ground truth. One of such works is from Bonechi \etal \cite{bonechi2019cocots}. As numerous datasets provide bounding–box level annotations for text detection, the authors obtained pixel-level text masks for scene images from the available bounding–boxes exploiting a weakly supervised algorithm. However, a dataset with annotated bounding boxes should be provided, and the bounding box approach is not suitable for unconstrained text.
Some few works that make pixel text segmentation in manga could be found on GitHub. One is called \enquote{Text Segmentation and Image Inpainting} by \texttt{yu45020} \cite{yu45020} and the other \enquote{SickZil-Machine} by U-Ram Ko \cite{Sickzil}\cite{Sickzil2}. Both attempt to generate a text mask in the first step via image segmentation and inpainting with such mask as a second step. In \textit{SickZil-Machine}, the author created pixel-level text masks of the  \texttt{Manga109} dataset, but has not publicly released the labeled dataset. The author neither released the source code of the method but has provided an executable program to run it. In \texttt{yu45020}'s work, the source code has been released.

\textbf{Pixel-level segmentation} 
Outside text, there are multiple works that do segmentation of different objects. One of such works is HRNet \cite{hrnet1} \cite{hrnet2}, which has top scores in many tasks, including segmentation with cityscapes dataset \cite{cityscapes}.


\section{Evaluation Metrics}\label{metrics}

Metrics such as recall, precision, and F-measure or F$_1$ score \cite{pf1} 
are widely used to evaluate binary segmentation models in images. 
These are defined as:
\begin{align}
Precision=\frac{TP}{TP+FP}
\label{precision}
\end{align}
\begin{align}
Recall=\frac{TP}{TP+FN}
\label{recall}
\end{align}
\begin{align}
 F \textendash measure = F_1 = Pixel \, F1= PF1 = \frac{2\ Recall\ Precision}{Recall + Precision}
 \label{pf1}
\end{align}
with TP being pixels that were correctly segmented as text (true positive), FP being pixels that were wrongly segmented as text (false positives) and FN being pixels that were wrongly segmented as background (false negatives). Other standard evaluation measures in document binarization are PSNR and DRD \cite{dibco2019}. PSNR is a similarity measure between two images. The higher the value of PSNR, the higher the similarity of the two images. 
The Distance Reciprocal Distortion Metric (DRD) has been
used to measure the visual distortion in binary document images \cite{drd}. The lower the DRD, the lower the distortion.

\begin{figure}[h]
\begin{center}
   \includegraphics[width=0.42\linewidth]{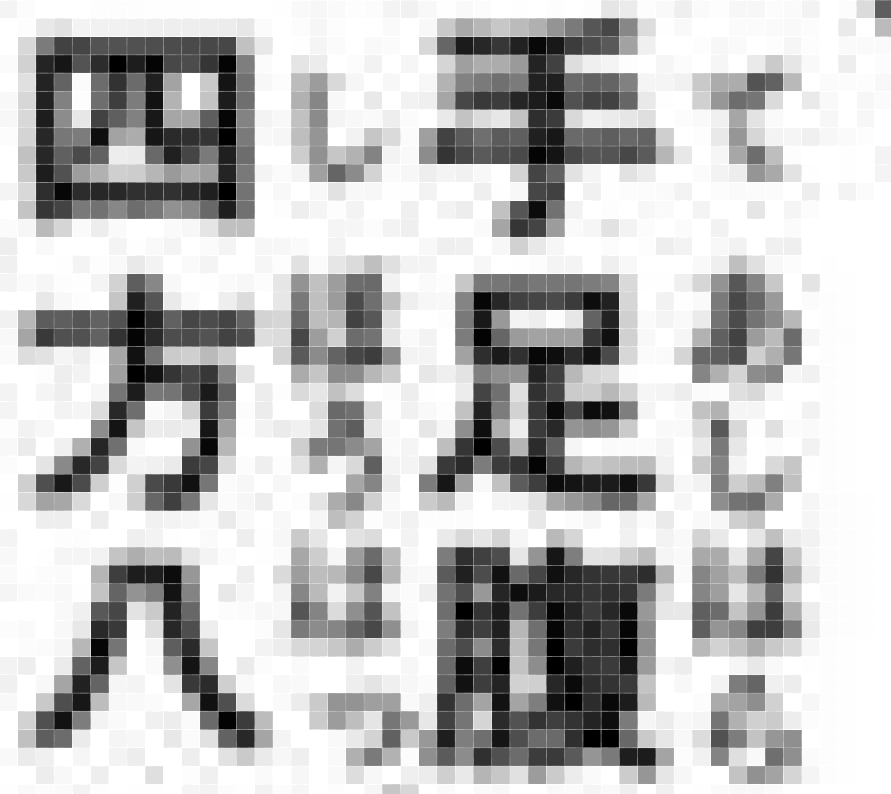}
\end{center}
   \caption{Example of text inside the speech bubble zoomed in. Note that text boundary is unclear and prone to error due to artifacts. 
   Image from ``Akkera Kanjinchou'' \textcopyright {Kobayashi Yuki},  \texttt{Manga109} dataset \cite{manga109}\cite{Matsui_2016}\cite{ogawa2018object}   
   }
\label{fig:blurring}
\end{figure}

Recall, precision and, F-measure metrics assume that the data is perfectly labeled and allow no compromises on the boundary, which is the part most prone to error. In many tasks, such as segmenting vehicles, this doesn’t matter much as the area of a car is very big compared to the area that might be wrongly labeled, so the human error in labeling won’t account much to influence metrics. With text, however, this is not the case. Not only are characters usually small, but also the boundary is many times unclear because of artifacts and blurring, as noted in Fig. \ref{fig:blurring}. Another issue is that a large text character can have the same area as 100 small characters, making a model that correctly matches most of its pixels but none of the other 100 characters, as good as one matching the 100 small ones but little of the big one.

\begin{figure}[h]
\centering
    \begin{subfigure}[b]{0.22\textwidth}
        \includegraphics[width=\textwidth]{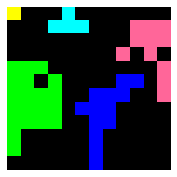}
        \caption{Ground Truth}
        \label{fig:watersheda}
    \end{subfigure}
    \begin{subfigure}[b]{0.22\textwidth}
        \includegraphics[width=\textwidth]{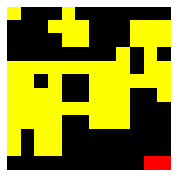}
        \caption{Prediction}
        \label{fp}
    \end{subfigure}
    \begin{subfigure}[b]{0.22\textwidth}
        \includegraphics[width=\textwidth]{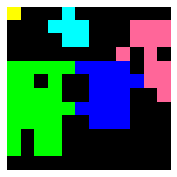}
        \caption{Watershed}
    \end{subfigure}    
   \caption{ Example of the watershed algorithm matching prediction pixels to ground truth connected components. (a) Masks of five ground truth connected components (labeled in different colors to be distinguished); (b) Predicted text segmentation mask. We marked in red (bottom, right) a predicted connected component with no correspondence to any ground truth connected component; (c) The predicted mask is matched with the ground truth using the watershed algorithm to obtain the evaluation metrics. Five detections matching ground truth connected components are obtained}
\label{fig:watershed}
\end{figure}

Calarasanu \etal \cite{calarasanu.phd}\cite{calarasanu.16.iwrr}\cite{calarasanu} have proposed several metrics to account for these issues. In this work, we have adopted an approach similar to theirs. In addition to the standard pixel metrics, we calculate metrics based on connected components. A connected component in these images is a region of adjacent pixels, considering its 8  neighbors, sharing the same value (see Fig. \ref{fig:watersheda}). 

Given a ground truth connected component $G_{i}$ and its matching detection $D_{j}$, its accuracy and coverage are defined as: 

\begin{align}
 Acc_i=\frac{Area(G_{i}\bigcap D_{j})}{Area(D_{j})} \label{eq:accuracy}\\
 Cov_i=\frac{Area(G_{i}\bigcap D_{j})}{Area(G_{i})} 
\end{align}

To account for multiple detections matching a single ground truth or a single detection matching multiple ground truths, we apply the watershed algorithm to match prediction pixels to a single ground truth, as seen in Fig. \ref{fig:watershed}. 

We define $tp$ as the number of ground truth connected components that have at least one pixel of detection associated with it. We define $fp$ as the number of detected connected components which had no correspondence to any ground truth (see Fig. \ref{fp}).

Given a dataset with $m$ ground truth connected components and $d$ detections, 
we define the following metrics: quantity recall $R_{quant}$, quantity precision $P_{quant}$, quality recall $R_{qual}$, quality precision $P_{qual}$ and F1 quality $F1_{qual}$: 

\begin{align}
 R_{quant}=\frac{tp}{m} \\
 P_{quant}=\frac{tp}{tp + fp}
 R_{qual}=\frac{\sum_{i=0}^{tp} Cov_i}{tp} \\
 P_{qual}=\frac{\sum_{i=0}^{tp} Acc_i}{tp} \\
 F1_{qual}= \frac{2\  R_{qual}\ P_{qual}}{ R_{qual} + P_{qual}} \label{f1qual}
\end{align}

Global recall $GR$, global precision $GP$,  and global F$_1$ $GF1$ are defined as:
\begin{align}
 GR=R_{quant} R_{qual}=\frac{\sum_{i=0}^{tp} Cov_i}{m} \\
 GP=P_{quant} P_{qual}=\frac{\sum_{i=0}^{tp} Acc_i}{tp + fp} \\
 GF1= \frac{2\ GR\ GP}{GR + GP}
\end{align}

\begin{figure}[h]
\centering
    \begin{subfigure}[b]{0.3\textwidth}
        \includegraphics[width=\textwidth]{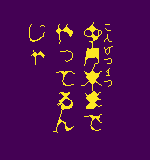}
        \caption{Eroded GT}
    \end{subfigure}
    \begin{subfigure}[b]{0.3\textwidth}
        \includegraphics[width=\textwidth]{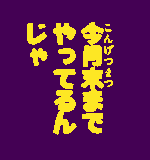}
        \caption{Original GT}
    \end{subfigure}
    \begin{subfigure}[b]{0.3\textwidth}
        \includegraphics[width=\textwidth]{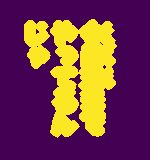}
        \caption{Dilated GT}
    \end{subfigure}    
   \caption{Example of segmentation masks used in normal mode and  relaxed mode metrics. (a) The eroded mask under-segments the ground truth mask. It is used for coverage in relaxed mode; (b) Ground truth mask. It is used in normal mode; (c) The dilated mask over-segments the ground truth. It is used for accuracy in relaxed mode. In relaxed mode, if the network predicts the ground truth of b), it has 100\% of accuracy. However, if the prediction is inside the dilated mask of c), it also has 100\% of accuracy. Accuracy is measured with Equation \ref{eq:accuracy}. 
   }
\label{fig:dilation_erosion}
\end{figure}

We calculate metrics in  normal and relaxed mode. Normal mode assumes that the dataset is perfectly labeled. Relaxed mode tries to lessen the effect of wrong boundary labeling (Fig. \ref{fig:dilation_erosion}). In normal mode, we calculate the metrics using the segmentation masks of the dataset without modification.  In relaxed mode, an eroded version of the ground truth is used to calculate coverage while a dilated version is used to calculate accuracy. In both modes, we consider there is no match to a ground truth component when there is no intersection between the eroded version and prediction, as the eroded version is the most relevant part to detect. For both erosion and dilation, a cross-shaped structuring element is used (connectivity=1).

\section{Dataset}\label{dataset}

There are very few datasets of images with text and their corresponding pixel level mask (Table \ref{table:datasets}). This is mainly due to the large amount of time required to label them properly.  
However, 
most of them correspond to real-world images, which differ greatly from manga, and lack useful context such as speech balloons. Besides, 
most text in these datasets is in English and manga is in Japanese.  Last, in these datasets, the text fonts are not hand-drawn in unrestricted orientations and with artistic styles like in manga. 

\begin{table}[h!]
\centering
 \caption{Datasets with pixel-level segmentation masks of text characters
 }
\begin{tabular}{ |c|c|c|c|c|c|c|}
 \hline
\textbf{Dataset}   & \textbf{Type}  & \textbf{Language} & \textbf{Script} & \textbf{Images} & \textbf{Orientation} \\
 \hline
\texttt{ICDAR 2013} Scene & Scene  & English & Latin & 462 & Horizontal\\ 
 Text Challenge \cite{icdar2013} &   &  &  &  & \\
\hline
\texttt{ICDAR 2013} Born & Digital  & English & Latin & 551 & Horizontal\\ 
Digital Challenge \cite{icdar2013} &   &  &  &  & \\ \hline
\texttt{Total-Text}   & Scene  & English & Latin & 1555 & Multi-oriented,  \\ 
  \texttt{(2017)}\cite{total_text}&  & &  &  & Curve  \\ \hline
\texttt{DIBCO} \cite{dibco_2019}  & Document  & European &  Latin & 136 & Horizontal   \\ 
\texttt{(2009-2019)} & &  languages & &  &\\ \hline
\texttt{KAIST}\cite{kaist} & Scene  & English, & Latin,Korean & 2483 & Horizontal\\ 
 &   & Korean &  &  & \\ \hline
Ours & Document   & Japanese & Kanji,Hiragana,  & 900  &Unrestricted \\
 &    &  &  Katakana,Latin &   & \\
\hline
\end{tabular}
\label{table:datasets}
\end{table}

\texttt{Manga109} \cite{manga109}\cite{Matsui_2016}  is the largest public manga dataset, providing bounding boxes for many types of objects, including text. However, it does not have pixel-level masks, and not all text has a bounding box. It is composed of 109 manga volumes drawn by professional manga artists in Japan, with all pages in black and white except for the first few pages of each volume. This is because unlike American and European comics which tend to be in color, manga is usually in black and white. 

Datasets of synthetic images could be made using manga-style images, such as the images of the \texttt{Danbooru2019} dataset \cite{danbooru2019}, selecting the images  without text, and adding text to them of a particular font and size. However, randomly adding text characters anywhere does not replicate where the text is naturally placed in manga, as much text is inside speech bubbles. 
Synthetically replicating the speech balloons is not easy either, as they are not always a simple rectangle or ellipse like shape. Besides, text outside speech balloons are part of the artwork, and usually have artistic styles of the author.

Taking into account all these issues, we decided to create our own dataset with pixel-level annotations. We chose to use images from \texttt{Manga109}, as it is a known public dataset, features a wide range of genres and styles, and the manga authors have granted permission to use and publish their works for academic research. To cover as many different styles as possible, few images from many manga volumes are preferable to a lot from few volumes, as long as those few are enough for the network to learn its style. After observing many examples, we concluded that the first ten images of each manga volume in the \texttt{Manga109} dataset were a suitable number, as that included the cover of the manga and a few pages of the actual content. Thus we manually annotated with pixel-level text masks
the first ten images from 45 different digital mangas, 
 totalizing 450 images.
As each manga image in the  \texttt{Manga109} dataset corresponds to 2 pages of a physical manga, we digitally annotated 900 physical pages of mangas.

\begin{figure}[h]
\centering
\begin{subfigure}[b]{0.45\textwidth}
   \includegraphics[width=\linewidth]{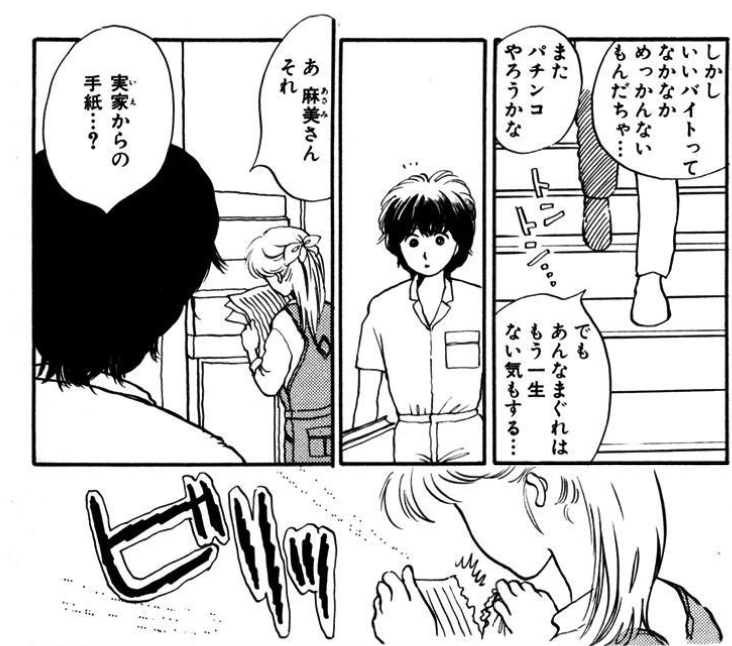}
            \caption{}
\end{subfigure}
\begin{subfigure}[b]{0.45\textwidth}
   \includegraphics[width=\linewidth]{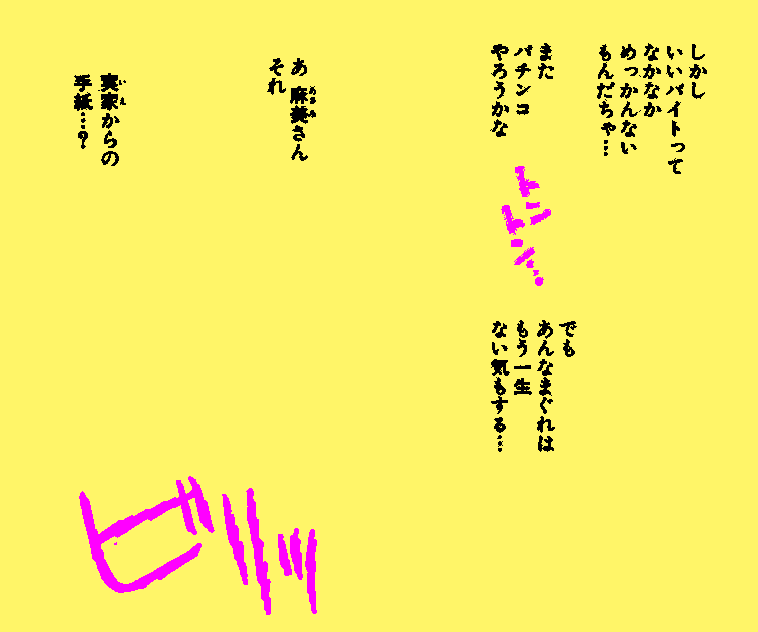}
            \caption{}
\end{subfigure}
   \caption{
   Example of segmentation mask in our dataset. (a) Original image. Speech text is usually inside balloons and sound effects outside. Note the sound effects, near the stairs, and ripped paper. 
   Image from ``Aisazu Niha Irarenai'' \textcopyright {Yoshi Masako},  \texttt{Manga109} dataset \cite{manga109}\cite{Matsui_2016}\cite{ogawa2018object}.
   (b) Corresponding segmentation mask in our dataset. The text inside speech balloons is considered as an easy detection task and labeled with black. Text outside balloons is considered a difficult text detection and labeled with pink. Non-text pixels are labeled with yellow
   }
\label{fig:example1}
\end{figure}

Instead of a simple binary mask (text and non-text), we label the dataset with 3 classes (Fig. \ref{fig:example1}, b): non-text, easy text (text inside speech balloons), and hard text (text outside speech balloons). While we still use the binary version for training, we use this separation of difficulties on text characters for a better understanding of model performance in metric evaluation.

In Table \ref{table:dataset_metrics} we show standard metrics for a U-Net model trained on diverse datasets and tested with our dataset. It is clear that the segmentation of text in manga is greatly improved if the model is trained with our dataset.

\begin{table}[ht!]
\centering
 \caption{Metrics on our dataset training with other datasets. All trained with resnet34 U-Net for 10 epochs, saving model when PF1 score improved on manga dataset. To test against our own dataset, K-fold was used over 5 folds. GF1 is based on connected components while the rest are all pixel based metrics. Precision refers to Eq. \ref{precision} and recall to Eq. \ref{recall}. Synthetic Text Danbooru2019 is a dataset that we built taking 75929 images without text from the Danbooru2019 \cite{danbooru2019} dataset and adding randomly Japanese characters from machine fonts.}
\begin{tabular}{ |c|c|c|c|c|c|c|  }
 \hline
\textbf{Dataset}  & \textbf{PF1} & \textbf{GF1}  & \textbf{Precision}  & \textbf{Recall} & \textbf{DRD} & \textbf{PSNR} \\
 \hline
\texttt{DIBCO}  & $26.32$ & $13.47$ & $19.09$ & $42.37$ & $74.71$ & $11.64$ \\
\texttt{ICDAR 2013} Scene Text Images  & $45.86$  & $40.47$ & $39.37$ & $54.91$ & $38.14$ & $14.53$ \\
\texttt{ICDAR 2013} Born-Digital Images   & $40.90$ & $39.67$ & $51.94$ & $33.73$ & $28.42$ & $16.22$ \\
\texttt{Total-Text}  & $42.49$ & $42.79$ & $47.10$ & $38.70$ & $29.19$ & $15.66$ \\
\texttt{KAIST}  & $45.69$ & $29.17$ & $44.55$ & $46.90$ & $32.64$ & $15.30$ \\
Synthetic Text \texttt{Danbooru2019}  & $57.03$ & $42.83$ & $76.01$ & $45.64$ & $18.88$ & $18.24$ \\
Ours   & $\bm{76.22}$ & $\bm{82.64}$ & $\bm{80.41}$ & $\bm{72.50}$ & $\bm{12.83}$ & $\bm{21.06}$ \\
  \hline
\end{tabular}
 \label{table:dataset_metrics}
\end{table}

\section{Methodology}\label{model}

Our text detector model employs a U-net  \cite{unet} architecture with a pre-trained resnet34 \cite{He2015DeepRL} backbone. Despite having been pre-trained with ImageNet, which features images quite different from  manga, it has proved to work well. We implemented the model in PyTorch. We used the \texttt{fastai} U-Net model \cite{fastaiunet} and trained the network with the \texttt{fastai} library \cite{fastai}\cite{fastaip}, making use of its one cycle policy, a modified version of the one initially devised by Leslie N. Smith \cite{smith}. The encoder part was frozen, and only the decoder part was trained, as the encoder already comes with the pre-trained weights from ImageNet. As we handle binary segmentation, a single channel is used as the last layer to provide the logits of a pixel being text. We later apply a sigmoid function and set 0.5 as a threshold to consider whether to classify it as text or background. As for the loss function, dice loss is used, which showed considerably better results than the simple binary cross-entropy loss.

The images of the  \texttt{Manga109} dataset are 1654 width and 1170 height. As they represent sheets of paper from physical books, in almost all cases (with some covers as the exception), the two pages from it have no text in the middle and can be split without affecting text characters. We took advantage of that and cut the images of our dataset in half, so we end up with 900 manga pages to train (see Section \ref{dataset}). The only data augmentation used is a 512x800 random crop for training. We tried a few other data augmentations such as flip and warp, but we didn’t notice any significant improvement.

We used K-Fold cross-validation with five folds to calculate all metrics, leaving 20\% as validation. Between transfer learning, one cycle policy, and a batch size of 4, results are obtained by training for ten epochs, which is completed in less than an hour on a single GeForce GTX 1080 Ti GPU for a single fold.

In the next section, we show how we used our metrics of Section \ref{metrics} to select an optimal loss function  and an optimal architecture for the model.

\section{Experiments}\label{experiments}

\subsection{Loss Function Selection}\label{loss}

Choosing an adequate loss function is a crucial step in the design of a machine learning model. 
We used our dataset to train a U-net network, which is a model commonly used in segmentation, with different loss functions. Then, we used the metrics of section \ref{metrics} to measure the performance of the model for each different loss function.

We trained a \texttt{fastai resnet18} U-net with each loss function  during ten epochs, with $0.001$ as the maximum learning rate. 
After trying different loss functions such as boundary loss, binary cross entropy and focal loss the highest scores in the metrics were obtained by $-log(DiceLoss)$ \cite{dice_loss}. Thus, this was the loss function that we choose for further experiments.

\subsection{Model Performance}
To create a good baseline, we made experiments with different methods and use the metrics to compare their performance for the binarization task of our dataset. The experimental results are shown in Table \ref{table:similar_works}.  

The \texttt{yu45020}'s xception \cite{yu45020}, the \texttt{yu45020}'s mobileNetV2  \cite{yu45020} and the \enquote{SickZil-Machine}  \cite{Sickzil}\cite{Sickzil2} are methods we found on Github that also aim to do pixel-level text segmentation in manga. We trained each one of the yu45020's architectures with our dataset for 10 epochs with all layers unfrozen. For SickZil-Machine the training code is not available so we used their executable to generate the predictions of Table \ref{table:similar_works}.

\begin{table}[h]
\centering
 \caption{Performance metrics  obtained  by similar methods under similar conditions for the task of binarizing  our manga dataset. All methods were trained with our \texttt{Manga109} labeled dataset except \textit{SickZil-Machine} because the author did not release the source code.  The results shown for \textit{SickZil-Machine} are for an executable program provided by the author. However, \textit{SickZil-Machine} was trained by his author with its own  \texttt{Manga109} dataset in which text was labeled at a pixel level. Precision refers to Eq. \ref{precision} and recall to Eq. \ref{recall}.
 }
\begin{tabular}{ |c|c|c|c|c|c|c|c|c|  }
 \hline
 \multirow{2}{*}{Author} & \multicolumn{6}{c|}{Normal} & \multicolumn{2}{c|}{Relaxed} \\
 \cline{2-9}
 & PF1 & GF1 & Precision & Recall & DRD & PSNR & PF1 & GF1 \\
 \hline
BCDU-net \cite{BCDUnet}  & $30.24$ & $15.08$ & $40.54$ & $27.32$ & $39.42$ & $14.80$ & $27.38$ & $16.19$ \\
\texttt{yu45020}'s xception   & $61.61$ & $62.48$ & $61.25$ & $62.09$ & $20.66$ & $17.77$ & $67.48$ & $79.35$ \\
\texttt{yu45020}'s mobileNetV2  & $47.97$ & $39.17$ & $49.43$ & $46.88$ & $28.12$ & $16.12$ & $50.54$ & $52.26$ \\
SickZil-Machine  & $52.07$ & $49.33$ & $41.77$ & $69.15$ & $35.75$  & $14.71$ & $64.66$ & $84.94$ \\
HRNet \cite{hrnet1}\cite{hrnet2} & $51.64$ & $51.34$ & $43.40$ & $63.85$ & $33.27$ & $15.40$ & $63.36$ & $74.41$ \\
\texttt{fastai} Resnet18 U-Net  & $75.82$ & $83.29$ & $75.53$ & $76.32$ & $14.06$ & $21.35$ & $76.87$ & $87.50$ \\
\texttt{fastai} Resnet34 U-Net & \bm{$79.36$} & \bm{$84.92$} &    \bm{$82.26$} & \bm{$76.71$} & \bm{$11.15$} & \bm{$21.91$} & \bm{$80.43$} & \bm{$89.26$} \\
 \hline
\end{tabular}
 \label{table:similar_works}
\end{table}

We also compared against BCDU-net \cite{BCDUnet}, a recent method for segmentation that has shown good results on multiple DIBCO datasets \cite{BCDUnetgithub}. BCDU-net is a deep auto-encoder-decoder network. This method applies bidirectional convolutional LSTM layers in U-net structure to non-linearly encode both semantic and high-resolution information with non-linearly technique. 
We chose the BCDU-net implementation that was based on DIBCO \cite{BCDUnetgithub} and trained it with our dataset in 5 folds, generating 40 random 128x128 patches from each image. Each fold was trained for 20 epochs.

HRNet \cite{hrnet1}\cite{hrnet2} is a very promising method for visual recognition with  state-of-the-art results in many  tasks and datasets. 
The authors show the
superiority of the HRNet in a wide range of applications, including human pose estimation, semantic segmentation, and
object detection. 
The configuration that we used was similar to the one provided for the cityscapes segmentation \cite{hrnet3}.  We trained each fold for 100 epochs each, changing the loss to $-log(DiceLoss)$.  

We also experimented with a lot of variations of the U-Net architecture provided with \texttt{fastai} library. As shown in Table \ref{table:similar_works}, the best binarization results were achieved by a \texttt{fastai} Resnet34 U-Net.  
For this network, we trained the decoder for 10 epochs, further trained the whole model for 5 epochs, and finally trained with whole images instead of random crops for 3 additional epochs. We kept this network as our baseline model because it achieved the top metric scores.

Last, we also experimented with 23 segmentation models from \texttt{qubvel}'s library \cite{qubvel}, training only the decoder and using the default parameters. However, the \texttt{fastai} U-Net network outperformed the models in this library in all metrics for this problem.

Our metrics allow us to find a simple and efficient model, that outperforms the other models 
especially on normal mode,  as shown in Table \ref{table:similar_works}. 

\begin{figure}[ht!]
\centering
\rotatebox{90}{\rlap{~Ground Truth}}
\begin{subfigure}[t]{0.3\textwidth}
    \includegraphics[width=\textwidth]{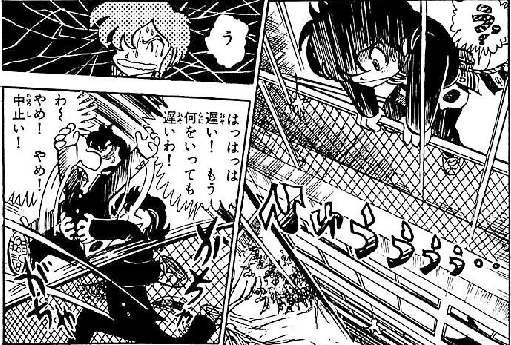}
    \caption{Manga image}
\end{subfigure}
\begin{subfigure}[t]{0.3\textwidth}
    \includegraphics[width=\textwidth]{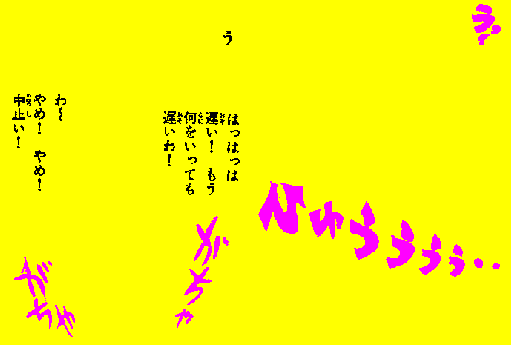}
    \caption{Ground Truth}
\end{subfigure}
\begin{subfigure}[t]{0.3\textwidth}
    \includegraphics[width=\textwidth]{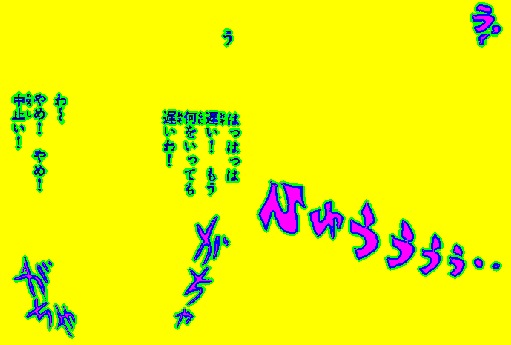}
   \caption{Relaxed GT}
\end{subfigure}

	\rotatebox{90}{\rlap{~Normal Mode}}
\begin{subfigure}[t]{0.3\textwidth}
    \includegraphics[width=\textwidth]{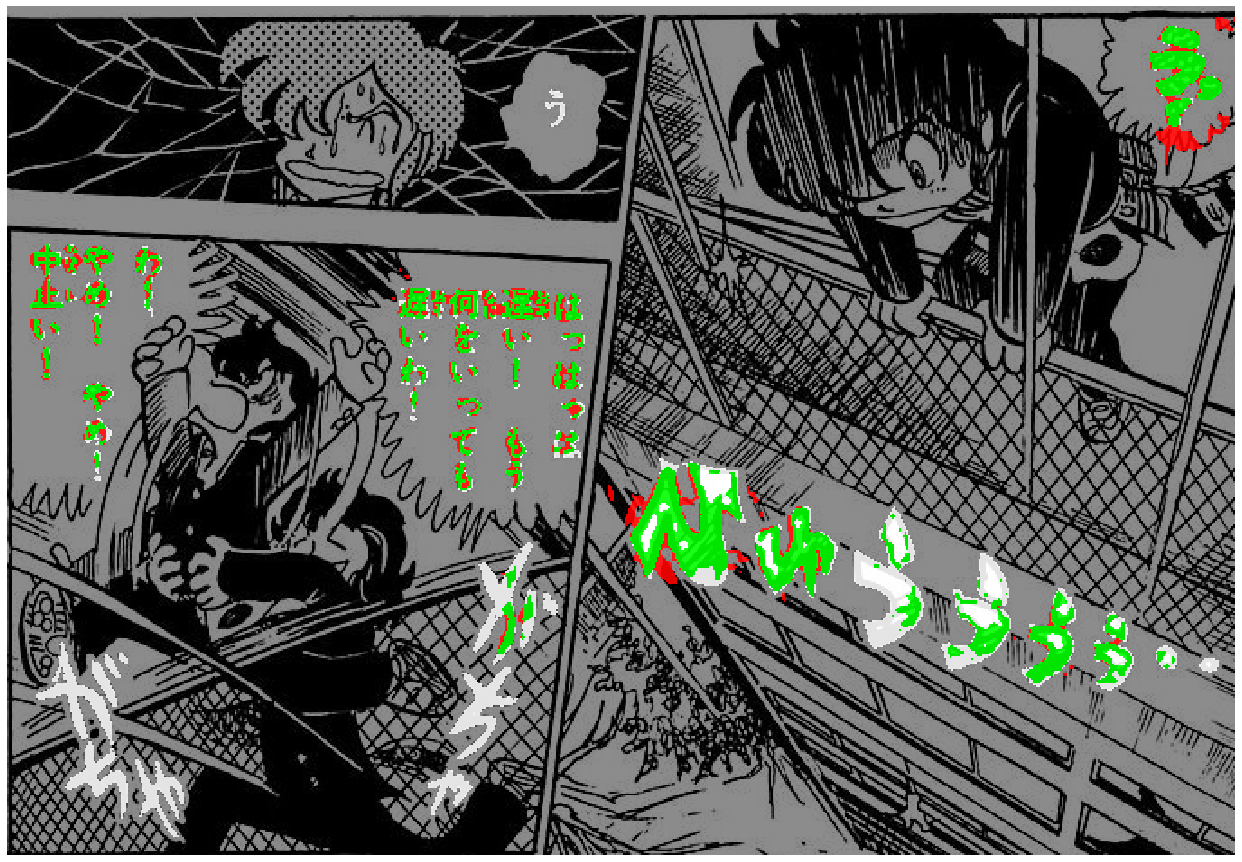}
    \caption{yu45020's xception}
    \label{fig:yumask1}
\end{subfigure}
\begin{subfigure}[t]{0.3\textwidth}
    \includegraphics[width=\textwidth]{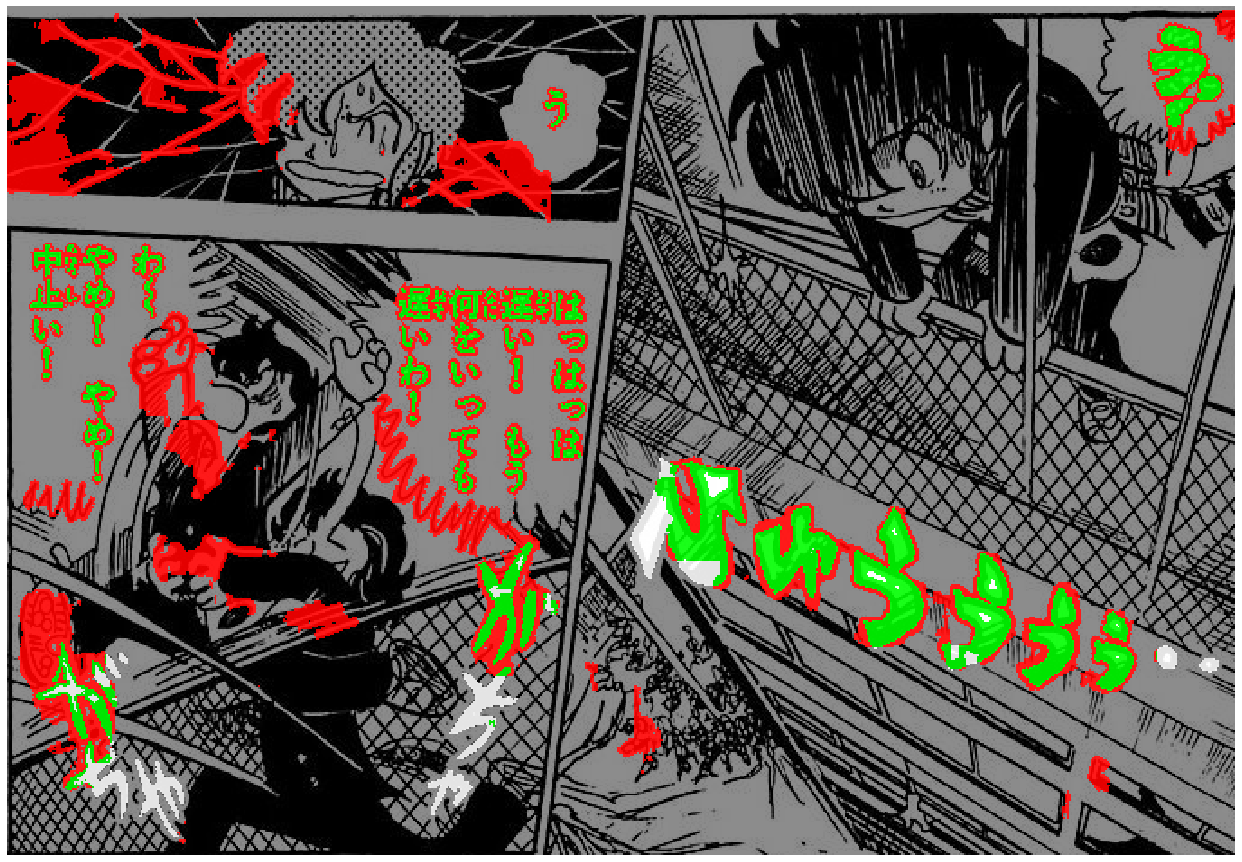}
   \caption{SickZil-Machine}
\label{fig:skmask1}
\end{subfigure}
\begin{subfigure}[t]{0.3\textwidth}
    \includegraphics[width=\textwidth]{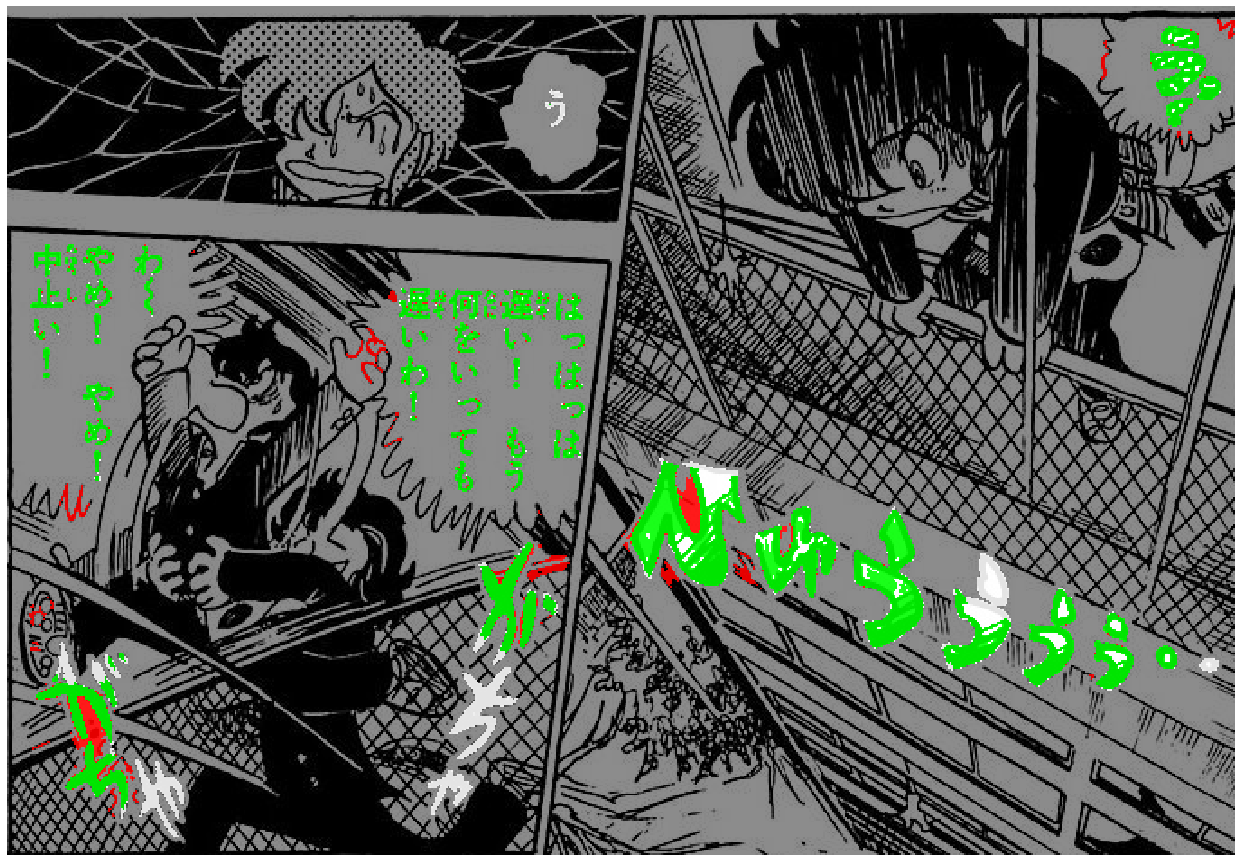}
   \caption{Ours}
    \label{fig:ourmask1}
\end{subfigure}

\rotatebox{90}{\rlap{~Relaxed Mode}}
\begin{subfigure}[t]{0.3\textwidth}
    \includegraphics[width=\textwidth]{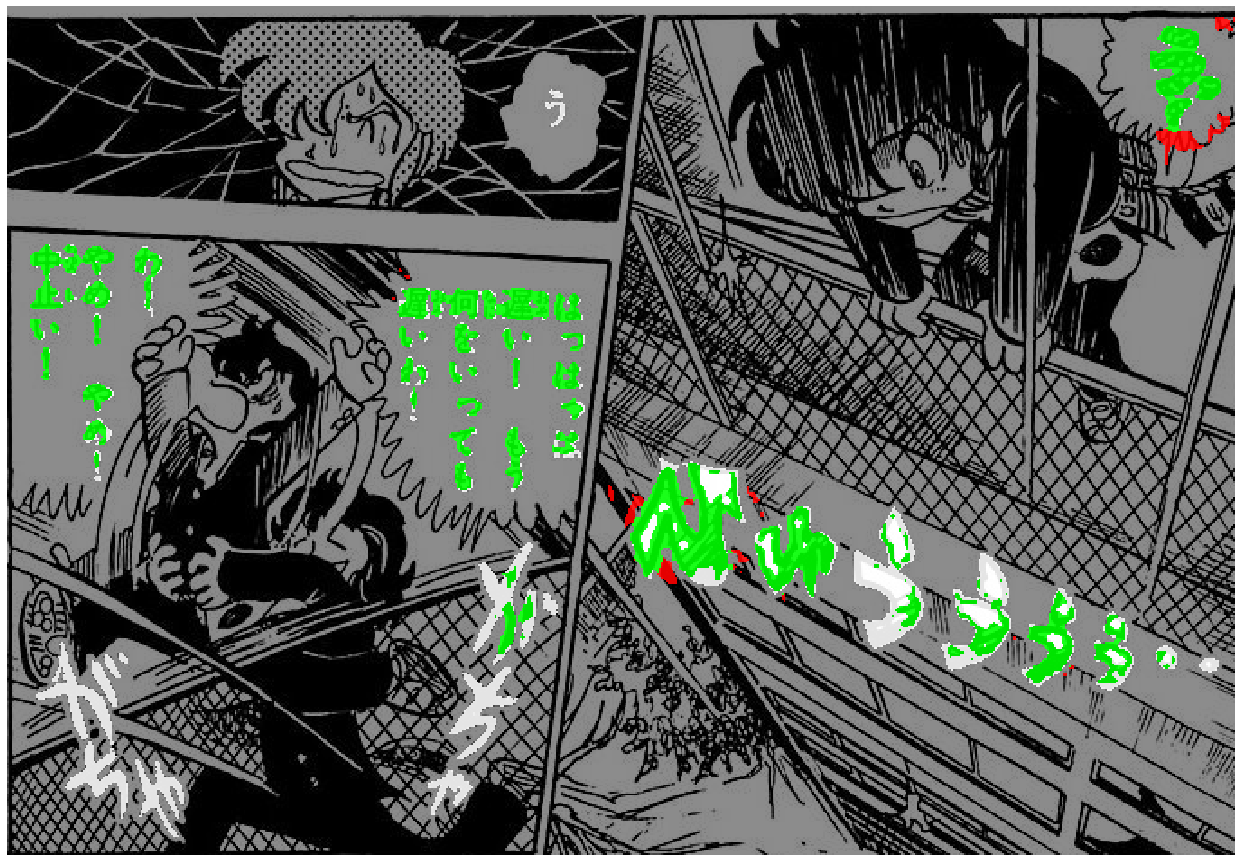}
    \caption{yu45020's xception}
    \label{fig:yumask2}
\end{subfigure}
\begin{subfigure}[t]{0.3\textwidth}
    \includegraphics[width=\textwidth]{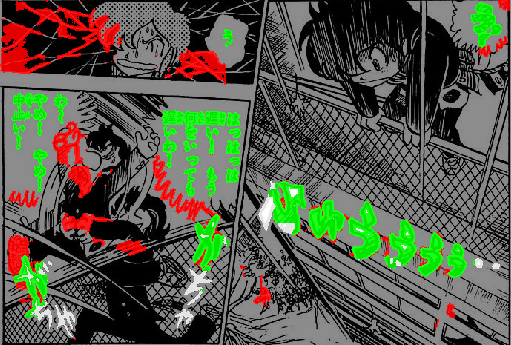}
   \caption{SickZil-Machine}
\label{fig:skmask2}
\end{subfigure}
\begin{subfigure}[t]{0.3\textwidth}
    \includegraphics[width=\textwidth]{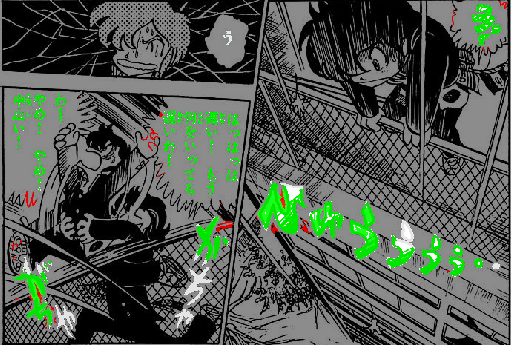}
   \caption{Ours}
   \label{fig:ourmask2}
\end{subfigure}
\caption{
Segmentation masks obtained by different methods. In red, false positives. In white, missing text. In green, text correctly segmented. (a) Image  extracted from ``BEMADER\_P'' \textcopyright {Hasegawa Yuichi},  \texttt{Manga109} dataset \cite{manga109}\cite{Matsui_2016}\cite{ogawa2018object}. (b) Ground Truth. In pink, hard text. In black, easy text. (c) Relaxed Ground Truth. In green, dilated area. In blue, ground truth pixels, that does not belong to the eroded mask.
(d, e, f) Normal mode results. (g, h, i)  Relaxed mode results. Note that in (h) boundaries between the small letters are lost, but they are still marked as true positives because the pixels are inside the relaxed dilated area}
\label{fig:similar_works_ours}
\end{figure}

 Fig. \ref{fig:similar_works_ours} shows an example of segmentation masks produced by the different methods. Our segmentation method misses some of the hard texts but has very few false positives (Figs. \ref{fig:ourmask1}, \ref{fig:ourmask2}). \textit{SickZil-Machine} covers some of those missing texts but also has much more false positives (Figs. \ref{fig:skmask1}, \ref{fig:skmask2}). 
 \texttt{yu45020}’s xception misses many of the hard texts, and detects the small letters with less precision than our model (Figs. \ref{fig:yumask1}, \ref{fig:yumask2}).

\begin{figure}[ht!]
\centering
    \includegraphics[width=\textwidth, height=0.4\textheight]{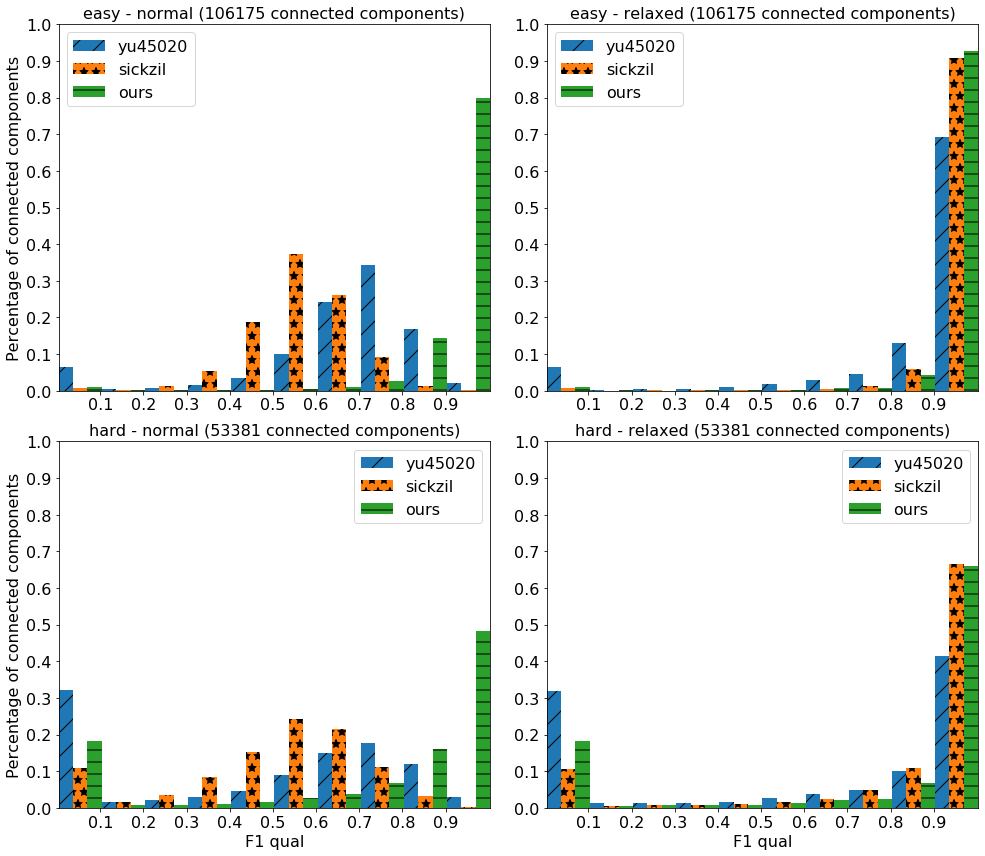}
    \caption{Histogram of F1$_{qual}$ (see Eq. \ref{f1qual}) of the different types of connected components. 
    The first row corresponds with easy text, and the second row corresponds with hard text. The first column corresponds with normal mode, and the second column corresponds with relaxed mode.
    For easy text, our method predicts most of the connected components with a high $F1_{qual}$ value. In normal mode our method has a much higher percentage of easy and hard connected components predicted with a high $F1_{qual}$ value than the other methods.
    }
\label{fig:similar_works_ours_histogram}
\end{figure}

For a global view of the performance on the different types of connected components and segmentation modes, we draw in Fig. \ref{fig:similar_works_ours_histogram} the histograms of $F1_{qual}$ (see Equation \ref{f1qual}). As our method fits the text characters without over-segmentation, it has less false positives, and our method clearly outperforms the other methods for $F1_{qual}$ in normal mode.
For the easy text case in relaxed mode, our method and \textit{SickZil-Machine} detect almost all the connected components.  Thus, we can see that there is little point in adding more data of easy text, as almost all easy components are detected with high F$_1$ score.

\section{Conclusions}

The detection and recognition of unconstrained text is an open problem in research.  
Standard methods developed for the Latin alphabet do not perform well with Japanese, due to Japanese having many more characters: about 2,800 common characters out of a total set of more than 50,000. Besides, each Japanese character is, on average, more complicated than an English letter.   
Japan is a country with an immense cultural heritage. Unfortunately, the complexity of the Japanese language constitutes a linguistic barrier for accessing its culture.  Automatic translation methods would contribute to overcome it.

In this work, we presented a study into unconstrained text segmentation at a pixel level in Japanese manga. We created a dataset manually annotating the text of manga images at a pixel level. Besides, we implemented special and standard metrics to evaluate the binarization task. We show that these tools, together with the  \texttt{fastai} library, allowed us to find a simple and efficient deep learning model that outperforms in most metrics previous works on the same task.  
Despite our focus in manga, the techniques proposed in this work could be expanded to do unconstrained Japanese text detection in other contexts. For instance, the text segmentation masks obtained by our method could be useful for Japanese OCR and inpainting in other Japanese graphical documents. 

The release of the dataset and metrics provided by this work would also enable other researchers and practitioners to find better models for this problem and compare results.  
To the best of our knowledge, this would be the first public dataset on text segmentation at a pixel level of Japanese manga. 

\section*{Code}

https://github.com/juvian/Manga-Text-Segmentation


\clearpage
%
%
\bibliography{egbib}
\bibliographystyle{splncs}
\end{document}